\pdfoutput=1

\documentclass[11pt]{article}

\usepackage[preprint]{acl}

\usepackage{times}
\usepackage{latexsym}
\usepackage[linesnumbered,ruled,vlined]{algorithm2e}
\usepackage{amsthm}
\usepackage{amsmath}
\theoremstyle{definition}
\newtheorem{definition}{Definition}
\usepackage{placeins}
\usepackage{wrapfig}

\usepackage[T1]{fontenc}

\usepackage[utf8]{inputenc}

\usepackage{microtype}

\usepackage{inconsolata}

\usepackage{graphicx}

\usepackage{booktabs}
\usepackage{multirow}
\usepackage{verbatim}
\usepackage{adjustbox}
\usepackage{tcolorbox}
\usepackage{subcaption}
\usepackage{amsfonts}
\usepackage{longtable}

\def \shortname{\textsc{UniPrompt}}
\def \uniprompt{\textsc{UniPrompt}}

\title{Task Facet Learning: A Structured Approach To Prompt Optimization}

\author{
  \textbf{Gurusha Juneja\textsuperscript{1,*}},
  \textbf{Gautam Jajoo\textsuperscript{2,*}},
  \textbf{Nagarajan Natarajan\textsuperscript{3}},
  \textbf{Hua Li\textsuperscript{4}},
  \textbf{Jian Jiao\textsuperscript{4}},
    \textbf{Amit Sharma\textsuperscript{3,\dag}}
  \\
  \\
  \textsuperscript{1}UC Santa Barbara,
  \textsuperscript{2}BITS Pilani,
  \textsuperscript{3}Microsoft Research,
  \textsuperscript{4}Microsoft Bing Ads
  \\
  \\
  \textsuperscript{*}These authors contributed equally to this work.
  \\
  \textsuperscript{\dag}Corresponding author: \href{mailto:amshar@microsoft.com}{amshar@microsoft.com}
}

\begin{document}
\maketitle
\footnotetext[5]{To appear in ACL Findings 2025.}
\begin{abstract}
 Given a task in the form of a basic description and its training examples, prompt optimization is the problem of synthesizing the given information into a text prompt for a large language model. Humans solve this problem by also considering the different facets that define a task (e.g., counter-examples, explanations, analogies) and including them in the prompt. However, it is unclear whether existing algorithmic approaches, based on iteratively editing a given prompt or automatically selecting  a few in-context examples,  can cover the multiple facets required to solve a complex task.  In this work, we view prompt optimization as that of learning multiple facets of a task from a set of training examples. We  exploit structure in the prompt optimization problem and break down a prompt into loosely coupled semantic sections. The proposed algorithm, \shortname{}, (1) clusters the input space and uses clustered batches so that each batch likely corresponds to a different facet of the task, and (2)  utilizes a feedback mechanism to propose adding, editing or deleting a section, which in turn is aggregated over a batch to capture generalizable facets.
 Empirical evaluation on multiple datasets and a real-world task shows that prompts generated using \shortname{} obtain higher accuracy than human-tuned prompts and those from state-of-the-art methods. In particular, our algorithm can generate long, complex prompts that existing methods are unable to generate. Code for \shortname{} is available at https://aka.ms/uniprompt. 
\end{abstract}

\section{Introduction}

\begin{figure*}
    \centering
    \includegraphics[width = 0.99\textwidth]{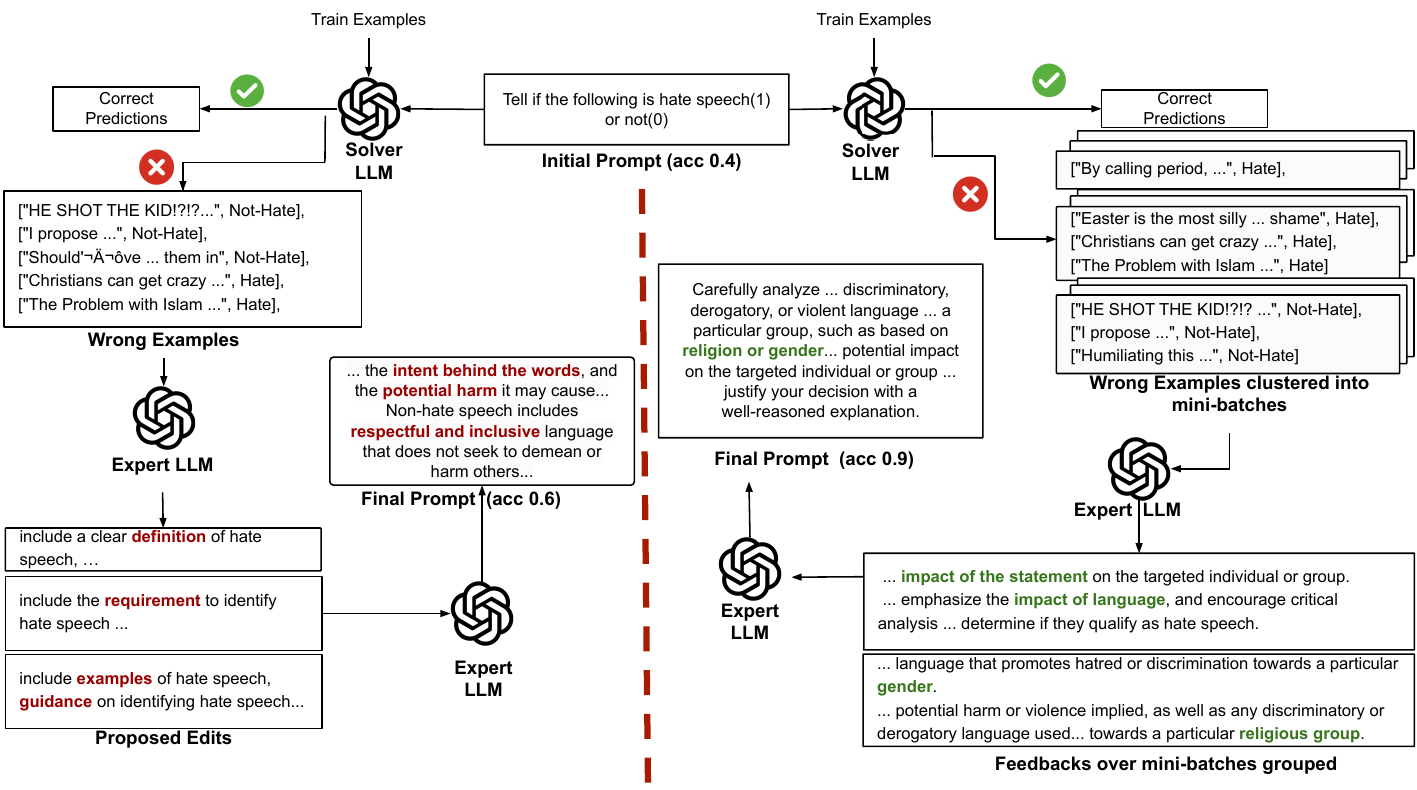}
    \caption{\textbf{Existing prompt optimization methods (left) versus \shortname{} (right) on the Ethos dataset}: [Left] State-of-the-art prompt optimization methods like \textbf{ProTeGi} ~\citep{pryzant2023automatic} sample from the questions wrongly answered by the current prompt, and use an expert LLM (e.g., GPT-4) to obtain feedback on the mistakes. This approach tends to give very general edits or overfits to specific examples.  [Right] In contrast, \shortname{} identifies key task \textit{facets} by: (1) clustering examples with similar task facets, and (2) employing a two-tier feedback-based update strategy. The resulting prompt updates extract generalizable concepts from the specific examples.}
    \label{fig:main}
\end{figure*}

 Given a task, choosing an input prompt is a key part of optimizing Large Language Model's (LLM) performance~\citep{cot,LLMO}. Minor changes in prompt can lead to performance gains or losses, necessitating prompt engineering~\citep{citeonpromptengg}. Typically, manually-developed prompts combine task description with a few in-context examples, along with modifiers like chain-of-thought~\citep{cot}. For greater accuracy, human prompt engineers spend considerable time to  identify errors with a current prompt,  consider the different facets of a task (e.g., counter-examples, explanations, analogies) that may fix those errors, and include them in the prompt. For instance, for a hate speech classification task, in addition to the definition, it may be helpful to specify the facets that lead to hate speech: the context of conversation, identifying intent, and differentiating hate speech from opinions or closely-related concepts such as vulgarity and profanity. \\ To avoid the above cumbersome manual process, recent work aims to automate the process of generating natural language prompts that are also interpretable. Since language tokens are discrete, this leads to a  challenging discrete optimization problem with a combinatorial space of possible outputs. Techniques for prompt optimization can be divided in two categories: \textit{non-directional}, e.g., random search~\citep{APE,autoinstructzhang} and genetic algorithms~\citep{LLMO,evoprompt}, where the sampling of new input is ``random'' and does not explicitly aim to reduce error on a train set; and \textit{directional}, where the sampling of new input depends on some error measure on a representative train sample. 
 Recently, more complex methods have been proposed in the second category including RL~\citep{tempera,RLPrompt}, updating prompts using feedback from auxiliary LLMs~\citep{evoke,pryzant2023automatic}, and optimizing the input to a small LM that generates the prompt~\citep{linuse2024,cheninstructzero}.
 While these techniques focus on editing parts of a given prompt, they are developed with the goal of obtaining a concise description of the task. None of these focus on ensuring multiple facets of a task are added to the prompt. \\ In this paper, we propose \uniprompt{}, a prompt optimization method to cover diverse, multiple facets of a task and  improve overall accuracy. To simulate the manual prompt engineering process, we propose that prompts be constructed from individual \textit{sections}, where each section may correspond to a different facet that humans may consider for the task. Prompt editing proceeds at a section-level: we can add, edit or delete a section from the prompt. 
Similar to \cite{pryzant2023automatic,evoke}, prompt edits are based on an auxiliary LLM's \textit{feedback} about example predictions with the current prompt. We contribute two key insights in this feedback-based optimization process. First,  we find that the feedback on a single example  or a randomly selected batch of examples does not yield generalizable facet descriptions. Instead, we propose clustering the inputs and creating mini-batches such that each mini-batch is sourced from a single cluster. Second, even with clustered batches, the feedback tends to overfit to specific examples or their properties. To generate a prompt edit that conveys a generalizable concept relevant to the task, we propose generating edits at a mini-batch level and then aggregating them at the batch level to yield the final edit (Figure \ref{fig:main}). 
While the two insights may appear \textit{simple}, we show that they significantly improve extracting diverse task facets. 

We evaluate \uniprompt{} on several benchmarks where it consistently achieves higher accuracy than existing prompt optimization methods.  On Ethos, a hate speech dataset, \uniprompt{} obtains 94\% accuracy whereas the next best method obtains 82\%. Even though UniPrompt focuses only on the instruction and does not include any in-context examples, we find that its instruction-only accuracy is often higher than methods such as DSPy~\citep{khattab2024dspy} that optimize both.  In the few-shot setting, we also compare \uniprompt{} to MedPrompt~\citep{medprompt}, a state-of-the-art prompt composition method. We find that \uniprompt{}, requiring only one LLM call at inference time,  obtains the same accuracy as MedPrompt that requires five calls. If we allow multiple calls to \uniprompt{}, we obtain over 4\% accuracy gains. Finally, we also evaluate \uniprompt{} on a real-world semantic matching task in a web search engine. Compared to the best manual prompt, the prompt generated from \uniprompt{} leads to over 5\% increase in accuracy on the rare class and nearly 2\% accuracy increase overall. 

\section{Related Work}

Here, we highlight relevant work that are not addressed in the manuscript otherwise. 
\citet{RLPrompt} present a discrete prompt optimization method, RLPrompt, using reinforcement learning, where a policy network learns to generate effective prompts through reward-based training, with an emphasis on enhancing training efficiency through effective reward stabilization techniques.
A drawback of such automatic prompt optimization approaches \citep{pryzant2023automatic,APE,RLPrompt,LLMO} is that the prompts generated tend to be short, often comprising only one or two sentences, which may not fully encapsulate the complexity of the task at hand. 

Another recent line of work leverages human feedback. Automated Prompt Optimization with Human Feedback ~\citep{lin2024prompt} optimizes prompts for black-box LLMs using human preference feedback. Besides the obvious overhead, it might also introduce potential biases. 

Prior research \citep{long2,long1} has highlighted the significance of specific sections within prompts. 
However, existing methods do not specifically target the optimization of individual sections and their respective contents within the prompts. \citet{longprompt} investigate the use of greedy and genetic algorithms to edit lengthy prompts. Their method focuses on paraphrasing one line at a time starting from an existing prompt, compared to our goal of learning facets of a task from scratch. Another orthogonal line of work explores algorithmic selection of in-context examples~\citep{ethos_cite_1,icl_rw_1, icl_rw_2, icl_rw_3, icl_rw_4}.
\section{UniPrompt: Capturing  Task Facets }
\label{sec:method}

State-of-the-art prompt optimization methods such as
ProTeGi \citep{pryzant2023automatic} and TextGrad \citep{houtextgrad} iteratively optimize the prompt for a given task. At a high-level, they proceed as follows: (1) start with an initial prompt and a training dataset of $\langle$question, answer$\rangle$ pairs for the task, (2) randomly sample from the questions wrongly answered by the current prompt to form a batch, (3) use an expert LLM to obtain feedback on the random batch, (4) apply the feedback to the prompt. This procedure is illustrated in Figure \ref{fig:main} [Left]. Our work is motivated by three key observations.\\
\textbf{1}. \textbf{Larger models are more amenable to prompt optimization}. We observe that the change in the objective function (i.e., loss on a validation set for a given prompt) per change in input is relatively more stable for larger models like GPT-4 (Figure~\ref{fig:motivation}) than for GPT-3.5 (analysis in Appendix \ref{observation_1}).\\
\textbf{2}. \textbf{Clustered-batching improves the quality of text gradients} (i.e., feedback), as against the standard random batching adopted in state-of-the-art prompt optimization methods (Section \ref{sec:perf}).\\
\textbf{3}. \textbf{Two-tier feedback helps learn generalizable facets}. Collecting feedback from an expert LLM over mini-batches, and then summarizing the individual feedback texts via a second step (Section \ref{sec:perf}) helps learn generalizable task concepts in prompts. 

The proposed method \uniprompt{}, in Figure \ref{fig:main} [Right], makes two contributions. First, we follow a two-tier setup of synthesizing feedback for a batch of training examples. We break up a batch into mini-batches, collect feedback on each of the mini-batches and then use a separate prompt to aggregate the different feedback texts into a generalizable concept. 
Second, to increase chances that a mini-batch corresponds to a coherent facet, we periodically (re)cluster the training data and ensure that each mini-batch consists of examples from the same cluster. 
\\ Algorithm~\ref{alg:prompt} receives as input a one-line task description and a train set $D_t$ of $N$ $\langle$question $q_i$, answer $a_i \rangle$ demonstrations. It extracts key concepts or facets relevant to the task and updates prompt sections using them, with the goal of increasing accuracy on the validation set $D_v$. We assume access to an ``expert LLM'' such as GPT-4. 

\subsection{Task facet learning using examples}
Extracting task-relevant concepts from a set of examples to refine a prompt is a complex problem comprising multiple steps. Given a set of incorrect predictions, one needs to analyze what went wrong in each prediction, form hypotheses, aggregate the hypotheses to identify specific concepts that are relevant for the task. Then, for each concept, one needs to attribute which facet/section of the current prompt needs to be edited to incorporate the concept.  
These operations are highly model-specific and are difficult to execute reliably. Therefore, we exclusively rely on an expert LLM. \\ First, we prompt the expert LLM to diagnose mistakes (\textit{feedback}) in each example given the answer and chain-of-thought reasoning produced by the solver LLM. Subsequently, we use this feedback to generate precise edits for the prompt that may fix the error. These individual edits are then aggregated over a mini-batch and  fed back into the same LLM, which then identifies a few major edits to be applied to the current prompt. To aid in identifying major edits that correspond to generalizable facets, we propose to cluster the examples as a preprocessing step and create clustered batches, such that each cluster shares some common facet of the task. 


\label{sec:clustering}
\subsubsection{Clustering for identifying facets}
We explore two approaches for clustering: \textit{topic-based clustering}, and \textit{feedback-based clustering}. \\ \textbf{Topic-Based Clustering.} Given a set of examples, we identify $l$ topics spanning the entire train set.  This type of clustering is motivated by the observation that solver LLM may make similar mistakes on examples from the same topic. Hence, for such examples, a common edit to the prompt could improve predictions for all the examples.  To obtain the clusters, the expert LLM is prompted (for prompt see Appendix \ref{app:clustering_prompts_1}) to provide a broad sub-topic $t_i$ for each question. Then the resultant list of sub-topics $\{t_1, t_2,\dots,t_N\}$ is again clustered into $k$ topics $\{t'_1, t'_2,\dots,t'_l\}$ by prompting the expert LLM. Based on this clustering, each example ${q_i, a_i}$ is assigned a cluster $t'_j$ corresponding to $t_i$. \\ 
\textbf{Feedback-Based Clustering.} Examples that receive similar feedback based on a prompt's predictions can help identify task facets. Consider a physics-based task where two examples from different topics  obtain the same feedback from the expert LLM to edit the ``\textit{Rules}'' section of the prompt to include the statement, ``\textit{Draw all forces on each body before writing the equations}''. We argue that such examples can be clustered. This type of clustering makes the broad edit identification step easier. To obtain the clusters, we first evaluate all training examples against the current best prompt and store the feedback $f_i$ from the expert LLM, corresponding to each incorrectly answered example ${q_i, a_i}$ (all the correctly answered questions form one cluster). We then prompt the expert LLM to cluster these feedbacks $\{f_1, f_2,\dots, f_N\}$ into $l$ clusters (see Appendix \ref{app:clustering_prompts_2}). For each cluster, we create a batch ${q_i, a_i}$ corresponding to feedbacks in that cluster.

\subsubsection{Obtaining generalizable prompt edits}
\textbf{Two-tier Feedback. } To encourage generalizable feedback from the expert LLM, we obtain feedback at two levels: mini-batch and batch. Given a batch (created using clustering discussed above), we break it up into mini-batches. \\ For each mini-batch $m$, we construct a prompt consisting of incorrectly-answered questions in $m$, the chain-of-thought produced by the solver LLM, their incorrect predictions and the ground-truth answers. We ask the expert to provide one feedback for the mini batch (prompt is provided in Appendix \ref{app:feedback_prompts}). The expert can suggest the following edits: add a section or subsection, delete a section or subsection, and edit a section or subsection. \\ Given the different edits for mini-batches within a batch $b$, we invoke the expert LLM again to summarize these edits into a single section update.  This combination ensures some degree of smoothness at every update which helps stabilize training. To make sure that the expert is able to generate generalizable edits,  we additionally provide a random set of incorrect examples that are not in the current batch and ask it to suggest an edit based on the existing edits that can correct the errors. As before, the class of edits allowed is the same. \\ \textbf{History for effective exploration.} To ensure comprehensive, non-repetitive exploration of prompts, we also provide the batch-level history of edits~\citep{evoke,LLMO} in the mini-batch-level prompt. 
History $H[b]$ is presented as $\{e_i, acc_i - acc_{i-1}\}$ where $e_i$ is the edit proposed at the $i^{th}$ update and $acc_i$ is the accuracy of the $i^{th}$ updated prompt (See Appendix \ref{app:feedback_prompts} for the prompt). 

\begin{algorithm*}
  \caption{\uniprompt{}} \label{alg:prompt}
  \KwIn{Train set $D_t$, validation set $D_v$, initial prompt for the task $p_0$, one-line task description $T$}
  \KwOut{Optimized prompt $P^*$ for the given task}  
   $C \leftarrow$ \texttt{cluster}($D_t, p_0$), initialize history $H \leftarrow \{\}$, and validation accuracies $V \leftarrow$ [];\\
  Initialize a beam of size 2 with the initial prompt: $p_1 \leftarrow p_0$ and $p_2 \leftarrow p_0$  \\ 
  \For{epoch $e$ and each cluster c in C}{
    
    \For{each batch $b \in \texttt{batches}($C$)$}{
   $ F \leftarrow []$\\
    \For{each mini-batch $m \in \texttt{mini-batches}($b$)$}{
    Evaluate the best prompt on mini-batch: $a_m \leftarrow$ \texttt{LLM}($m,p_1$)\\
    Get expert feedback: $f \leftarrow$ \texttt{Feedback}($T$, $a_m$, $H[m]$)\\
    $F$.\texttt{insert}($f$)\\
    }
     Combine feedbacks over a batch: $F_b \leftarrow$ \texttt{Combine}($F$)\\
     Apply feedback to get updated prompts: $q_1 \leftarrow$ \texttt{apply}($F_b$, $p_1$); $q_2 \leftarrow$ \texttt{apply}($F_b$, $p_2$)\\     
     Update the beam: \textbf{if} not($p_1$ = $p_0$) \textbf{then} $p_2 \leftarrow$ \texttt{second-high-acc}([$p_1$, $p_2$, $q_1$, $q_2$], b) \\
      $p_1 \leftarrow$ \texttt{highest-acc}([$p_1$, $q_1$, $q_2$], b)
    }
    Evaluate the best prompt on validation set: $acc_v \leftarrow$ \texttt{evaluate($p_1$, $D_v$)}\\
    $V \leftarrow V$.\texttt{append}($acc_v$)\\
    \textbf{if} \texttt{early-stop-criteria} ($V$) \textbf{then} break \\
    \textbf{if} \texttt{recluster}($e$) \textbf{then}  $C \leftarrow$ \texttt{cluster}($D_t, p_1$)
  } 
  \Return{$p_1$ as $P^*$}\;  
\end{algorithm*}

\subsubsection{Editing the prompt}
\label{sec:updaterule}
 Once the final set of edits is received for a batch, we use the expert LLM to apply edits to the current prompt  (See Appendix \ref{app:editing_prompts} for the prompt). An edit is accepted only if it increases the validation accuracy (\textbf{Greedy}).  Alternatively, we maintain a beam of 2 best performing prompts based on validation accuracy, apply the edit to the two prompts, and update the beam to retain the top 2 performing prompts (\textbf{Beam}). To avoid overfitting on the train examples (or adding unnecessary information to the prompt), we employ early stopping in the optimization process (more details in Section \ref{sec:implementation}).


\subsection{Prompt Initialization}
\label{sec:init}
We use two types of initialization: \textbf{(1)} task description, i.e., $p_0$ has a single section titled \textit{Introduction} containing the input task description.  \textbf{(2)} finetune Llama2-13B model to generate a prompt with sections such as \textit{Introduction}, \textit{Tricks}, and \textit{Corner Cases}, similar to the initial prompt that a human prompt engineer may produce. To finetune, we use GPT-4 generated data consisting of (task description, section title, section contents) triples. Details and examples are in Appendices \ref{app:SLM_train_details} and \ref{app:initalization_example}.

\paragraph{Computational Complexity:} The complexity of clustering and of getting mini-batch and batch-level feedbacks per epoch is $O(N)$ expert LLM queries, where $N$ is the number of training examples. Details are in Appendix \ref{sec:complexity}.

\section{Experiments Setup}
\label{sec:setup}

\textbf{Datasets: }
\label{sec:datasets}
We perform comprehensive evaluation on five standard datasets:  
(1) Ethos \citep{ethos}, (2) ARC \citep{arc}, (3) MedQA \citep{medqa}, (4) GSM8K \citep{gsm8k}, and (5) BBH \citep{suzgun2022challengingbigbenchtaskschainofthought}.  Ethos, ARC, and MedQA contain multiple choice questions, and GSM8K contains questions with integer answers. BBH is a subset of 10 tasks, spanning 4 main categories, from the challenging BIG-Bench benchmark that requires multi-step reasoning. In addition, we also evaluate \uniprompt{} on the medical QnA datasets used in the MedPrompt \citep{medprompt} work; as well as two popular code generation datasets, HumanEval \citep{chen2021codex} and MBPP \citep{austin2021program}.  \\ 
\textbf{Implementation details: }
\label{sec:implementation}
We set the initial prompt $p_0$ for each task as the one-line task description. We use 200 examples as the train set, 200 examples as the test set, and 100 examples as the validation set for all the compared methods.  We use GPT-3.5-Turbo as the solver model. For \texttt{Feedback} and \texttt{Combine} in \uniprompt{}, we use GPT-4 as the expert (see ablation in Section \ref{sec:ablation}). 
We maintain a beam size of 2. Mini-batch sizes (and batch sizes) are constrained by the context length of GPT-4. We find that mini-batch sizes 3 to 5 and batch sizes 5 to 7 work the best for our datasets. The temperature of the LLMs for our method is set to 0 for reproducibility. 
We employ early stopping at batch-level in \uniprompt.  \\ \textbf{Baselines: }
\label{sec:baselines}
We compare \uniprompt{} with the following techniques and baselines: (1) \textbf{Task Description}: prompt is the one line task description that we use to initialize \uniprompt{}; (2) \textbf{Chain-Of-Thought} (or CoT) prompting \citep{cot}; (3) \textbf{Expert Prompt}: the prompt optimized by humans taken from prior works \citep{medprompt}; (4) \textbf{OPRO} \citep{LLMO}, that uses LLMs for discrete optimization over text prompts; (5) \textbf{ProTeGi} \citep{pryzant2023automatic} that proposes textual gradients and selects edits to prompts using bandit techniques; (6) \textbf{Evoke} \citep{evoke} that uses two instances of LLM, one that scores the current prompt, and the other that edits the prompt; (7) \textbf{EvoPrompt} \citep{evoprompt} that uses genetic algorithms to search through the space of prompts; (8) \textbf{TextGrad} \citep{houtextgrad}, state-of-the-art framework for automatic differentiation of prompts via text; (9) \textbf{DSPy} \citep{khattab2024dspy}, a recent programming model for optimizing LLM prompts; and (10) \textbf{MedPrompt} \citep{medprompt}, a state-of-the-art prompt composition method. 


\section{Results and Analysis}
\label{sec:results}
We present detailed quantitative and qualitative results, along with key ablations. 

\subsection{Performance of \shortname{}}
\label{sec:perf}
We start with the zero-shot setting, where we do not include labeled examples in the prompt for any of the compared methods. We report results for two versions of our method in Table \ref{tab:performance}, which differ in 
the combining strategy (from Section \ref{sec:updaterule})---beam search vs greedy. \\ \shortname{} variants significantly outperform the baselines including CoT and the state-of-the-art prompt optimization techniques like ProTeGi that crucially leverage LLMs for performing iterative prompt edits. \shortname{} is the best performing method on three out of four datasets. It achieves maximum gains on the Ethos dataset with a \(18.2\%\) increase in accuracy over the expert prompt. Further, we see accuracy increases of \(4.0\%\) on MedQA, \(3.5\%\) on GSM8k, and \(7.6\%\) on ARC-Challenge datasets. We show \uniprompt{} training behavior in Appendix \ref{app:trainingcurves}. \\ We also present comparisons to state-of-the-art DSPy method in the few-shot setting (8 \texttt{bootstrapped\_demos}) using two optimization settings provided by their framework. The last two rows of Table \ref{tab:performance} show that \shortname{} in the zero-shot setting convincingly outperforms DSPy in the few-shot setting, on three out of four datasets. \\ 
\begin{table*}[h]
  \centering
      \caption{Test accuracies (\%) of the compared methods with GPT-3.5-Turbo as the solver model in the zero-shot setting (\textbf{best} in bold; \underline{second best} underlined). The two \uniprompt{} rows are our proposed method. We  compare with few-shot methods in the last two rows (DSPy variants); *\textbf{best} in bold to distinguish the few-shot setting.}
    \begin{tabular}{l|r|r|r|r}
    \toprule
      Method    & Ethos & ARC & MedQA & GSM8K \\
    \toprule
    Task Description & 76.8 &79.7&52.7& 59.4\\
    Expert Prompt & 74.1 & 78.4& 53.1 &78.9\\    
    Llama Prompt (Section \ref{sec:init}) & 74.0 & 89.7 & 52.6 & 79.5\\
    \hline
   
    CoT & 72.0& 79.4&  50.3& 76.3\\
    OPRO &65.4 & 79.1& 53.3 & 77.1\\
    ProTeGi & 76.0& 78.8 &52.9 & 77.3\\
    Evoke & 63.5 & 89.0& 52.8 & 81.0\\
    EvoPrompt & 81.6 & \underline{89.9} & 50.3 & 81.4\\

    DSPy (\texttt{MIPRO v2}, zero-shot) & 79.7 & 82.8 & \textbf{61.9} & 77.3 \\ 
    TextGrad & 79.5 & 76.5 & 50.6 & 81.6 \\
    \hline
    \uniprompt{} (Init = Task Description) + Beam & \underline{92.3} & 86.0  & \underline{57.1}  & \textbf{82.4} \\
    \uniprompt{} (Init = Task Description) + Greedy    & \textbf{93.7}  & \textbf{90.5}  & 55.5  & \underline{82.3} \\
     \hline \hline
  DSPy (\texttt{BootstrapFewShotWithRandomSearch}) & 86.6 & 87.5 & *\textbf{68.5} & 74.3 \\
    DSPy (\texttt{MIPRO v2}, few-shot) & 84.0 & 86.0 &  62.9 & 79.7 \\
      
    \toprule
    \end{tabular}%
  \label{tab:performance}%
\end{table*}%
\textbf{Qualitative Analysis:} ProTeGi and TextGrad also adopt batching by randomly sampling from training examples where the solver LLM made mistakes. In the early iterations of optimization, there can be many such examples. So, do our key observations and hypotheses (beginning of Section \ref{sec:method}) hold empirically? We give some evidence below. \\ 1. \textbf{Employing clustering to create batches} (Section~\ref{sec:clustering}): An example feedback obtained on the Ethos dataset using \shortname{} is shown below:

\begin{quote}
\textit{The instruction should include..potential harm or violence implied, as well as any discriminatory or derogatory language used...towards a particular religious group. }

\textit{The instruction should include ...think about the impact of the statement on the targeted individual or group.}

\textit{The instruction should...language that prompts hatred or discrimination towards a particular gender.}
\end{quote} 
To contrast, we employ random batching as in the standard prompt optimization techniques, on the same dataset. The feedback obtained, given below, is relevant for the task, but fails to identify specific concepts.

\begin{quote}
\textit{The instruction should include a clear definition of hate speech...}

\textit{The instruction should include examples of hate speech, guidance on identifying hate speech...}
\end{quote}

The former feedback (\uniprompt{}) captures the facet of measuring impact on the targeted entity whereas the latter only captures religious and harm-based aspect of hate speech. \textbf{The same expert LLM is able to identify different facets due to clustered batches}. \\ 2. \textbf{Employing two-tier feedback}: In Section \ref{sec:method}, we argued that employing two-tier feedback strategy to aggregate the feedback texts encourages the expert LLM to propose edits that are generalizable. The following feedback is received on the Ethos dataset after the aggregation:

\begin{quote}
\textit{The instruction should...consider whether the statement contains discriminatory, derogatory, or violent language that promotes hatred or harm towards a particular group, such as based on religion and gender.}
\end{quote}
We see that two-tier feedback helps in distilling important aspects of the task implicit in the examples, rather than directly using or rephrasing the (limited) examples. \\ \textbf{Results on BBH:} From Table \ref{tab:bbh}, it is evident that \uniprompt{} shows a significant improvement over OPRO (that also evaluates on these tasks in their paper) for a majority of tasks. It achieves significantly higher accuracy in Boolean Expression (92.37\% vs. 78.74\%), Date Understanding (81.96\% vs. 52.59\%), and Navigate (77.16\% vs. 51.74\%). 

\begin{table}[h]
  \centering
    \caption{Test accuracies (\%) on BBH dataset with GPT-3.5-Turbo as the solver model}
    \centering
    \begin{tabular}{l|r|r|r}
      \toprule
      Task & Init & OPRO & \uniprompt{} \\
      \toprule
      \multicolumn{4}{l}{\textbf{Algo \& Multi-Step Arithmetic Reasoning}} \\
      \midrule
      Bool Exp. & 83.64 & 78.74 & \textbf{92.37} \\
      Logical Ded. & 29.53 & 38.97 & \textbf{39.62} \\
      Navigate & 60.95 & 51.74 & \textbf{77.16} \\
      \midrule
      \multicolumn{4}{l}{\textbf{Natural Language Understanding}} \\
      \midrule
      Snarks & 67.00 & 67.88 & \textbf{74.30} \\
      Disamb. QA & 53.30 & 57.43 & \textbf{67.05} \\
      Fallacies & 57.60 & 53.14 & \textbf{57.90} \\
      \midrule
      \multicolumn{4}{l}{\textbf{Use of World Knowledge}} \\
      \midrule
      Causal Judg. & 54.29 & 57.24 & \textbf{59.37} \\
      Movie Rec. & 58.04 & \textbf{77.81} & 71.80 \\
      Dates & 74.21 & 52.59 & \textbf{81.96} \\
      \midrule
      \multicolumn{4}{l}{\textbf{Multilingual Knowledge \& Reasoning}} \\
      \midrule
      Salient Trans. & 42.59 & 50.61 & \textbf{50.77} \\
      \bottomrule
    \end{tabular}%
    \label{tab:bbh}%
\end{table}

\subsection{Comparison with MedPrompt}
\label{sec:medprompt}
MedPrompt \citep{medprompt} is a recent, competitive prompting technique without any training component. It employs three key ingredients: (1) few-shot prompting, where five relevant examples are selected using k-nearest neighbors (kNN); (2) CoT reasoning on the selected examples; and (3) self-consistency and ensembling with option shuffling at inference time. They evaluate on 4 medical datasets (that none of the competing methods in Table \ref{tab:performance} evaluate on) using GPT-4 as the solver model. So, we compare \uniprompt{} in the same setting in Table \ref{tab:medprompt} (in Appendix \ref{additional_results}). \shortname{} (first row), which requires only one call at inference time, performs almost as well as MedPrompt (last row), which requires five calls, on three out of four datasets. As we incrementally add kNN few-shot, CoT, and ensembling to our prompt, we see a significant increase in accuracy of \(4.35\%\) on average across all datasets.

\subsection{Performance on generation tasks}
\label{sec:coding}
Our evaluations so far have been on multiple-choice QnA, math, and classification datasets. We now evaluate \shortname{} on generating code given a natural language specification. We use HumanEval \citep{chen2021codex} and MBPP \citep{austin2021program} datasets consisting of Python coding problems. We initialize with a simple prompt, ``\textit{You are a software engineer. You are given a function signature and a description of the function. You have to complete the function.}'' We use GPT-4-Turbo as both the solver and the expert LLM. \\ HumanEval does not have train or validation sets. So, we take random 100 examples from MBPP as train. Similarly, for MBPP, we take random 50 examples from HumanEval as train. We evaluate the final prompts on HumanEval and MBPP test sets. The results are given in Table \ref{tab:code} (in Appendix \ref{additional_results}). The metric is \% of solved coding problems (evaluated using the provided test cases) in the datasets. The prompts produced by \shortname{} outperform standard prompting of LLMs. 



\subsection{Results on a Real-world Task}
\label{sec:queryintent}
The task of inferring if two search queries share identical intent or not arises in search and recommendation pipelines. It is challenging because it requires domain knowledge (e.g., brands and product categories), and depends on cultural and geographical biases (e.g., ``cricket'' in UK vs. ``cricket game'' in the US). So, examples are crucial for engineering a prompt that generalizes well. \\ 
We sample 200 train, 50 validation, and a separate 2527 test user queries from a real proprietary application. More details on the dataset and the one-line initial prompt are  provided in Appendix \ref{app:dataset_real_world_task}. \\ 
The prompt obtained using \uniprompt{} improves over the best manual prompt by 5.77\% on the negative (rare) class, by 0.23\% on the positive class, and by 1.86\% overall on the test set. The learnt prompt captures the following task facets: (1) recognizing variations in names and abbreviations, and how they do not change the context; (2) recognizing brand specificity, and how even minor variations do change the context; and (3) recognizing the specificity of terms in queries, and how lack of specific terms can indicate departure of intent.

\subsection{Ablations}
\label{sec:ablation}

\textbf{Impact of Clustering, Inclusion of History, and Greedy Update:}
The results are shown in Table \ref{tab:ablation} (in Appendix \ref{additional_results}). We see that clustering as well as edit history components (Section \ref{sec:clustering}) are critical for performance of \shortname{} in all the datasets. We see a major drop of \(14.8\%\) in accuracy in the Ethos dataset when clustering is removed, and a  \(4.3\%\) drop when history component is removed. In all the datasets except GSM8K, we find clustering is more important than history. This can attributed to limited variability of question types (all grade-8 arithmetic) in GSM8K than in others. \\ We also find that the greedy update rule (Section \ref{sec:updaterule}) proves to be superior or competitive compared to beam search in relatively easier datasets --- where even less exploration produces good results, greedy proves to be a more effective update rule. On the other hand, in more complex datasets like MedQA, greedy appears to be a bad strategy. We also see that clustering examples based on feedback (``Fb Clustering'') is a better strategy than clustering based on topics, except for the Ethos dataset. 
\\ \textbf{Impact of Mini-batch size and Number of Clusters: }
\begin{table}[t]
  \centering
  \caption{Impact of \shortname{}'s key hyperparameters}
  \begin{tabular}{l|r|r}
    \toprule
    \textbf{Hyperparameter} & \textbf{Value} & \textbf{Accuracy} \\
    \midrule
    \centering
    \multirow{3}{*}{Mini-batch Size} & 2 & 84.90\% \\
                                     & 5 & 90.36\% \\
                                     & 8 & 91.15\% \\
    \midrule
    \multirow{3}{*}{Number of Clusters} & 2 & 85.81\% \\
                                        & 5 & 90.36\% \\
                                        & 10 & 87.82\% \\
    \bottomrule
  \end{tabular}
  \label{tab:impact_mini_batch_clusters}
\end{table} 
First, we vary mini-batch size in \uniprompt{}. The results for the ARC dataset are shown in Table \ref{tab:impact_mini_batch_clusters} (in Appendix \ref{additional_results}). With an increase in mini-batch size, we observe an increase in accuracy. That said, it is a hyperparameter, hence there will be an optimal number for each dataset. The mini-batch size affects the feedback based on the wrong examples that are obtained in each round. Next, we vary the number of clusters in \uniprompt{}. We find that the parameter has a clear impact on performance. We use the default choice of 5 clusters in all our experiments, which provides concise and generalizable feedback. 
\\ 
\textbf{Impact of initial prompt and Expert model capacity:} In Table \ref{tab:initprompt} (Appendix \ref{additional_results}), we find that one-line task description initialization for \uniprompt{} achieves the best accuracy on three out of four datasets. On ARC, initializing with the prompt generated by the Llama2-13B model gives significant improvement over other initializations. 
In Table \ref{tab:expertchoice} (Appendix \ref{additional_results}), we show \uniprompt{} improves prompts for more capable solver LLMs while using less capable expert LLMs.


\section{Conclusions}
\label{sec:limitations_and_future_work}
We presented a method inspired by the human prompt engineering process to generate complex prompts from scratch that include different facets of a task. Our algorithm provides significant improvements over baseline prompt generation methods on multiple standard datasets. Just like in-context learning \citep{ji2024submodular}, task facet learning could also benefit from connections to submodular optimization \citep{krause2014submodular}. We leave this as future work. 

\section{Limitations}
We provide an analysis of the impact of model size on the amenability to prompt optimization in Appendix \ref{observation_1}. However, in our evaluation, we only use GPT-3.5 (and in some cases GPT-4) as the solver LLM. We want to leave extensive evaluations of using open-source LLMs as solver LLMs, and perhaps even as expert LLMs, to future work. Further, we also want to evaluate on other generative tasks, besides the code generation task we study in the paper. 

\bibliography{custom}
\appendix

\appendix
\section{Appendix}
\subsection{When is directional text optimization feasible?} 
\label{observation_1}
\begin{figure*}[htbp]
    \centering
    \begin{subfigure}[b]{0.325\textwidth}
        \centering
        \includegraphics[width = \linewidth]{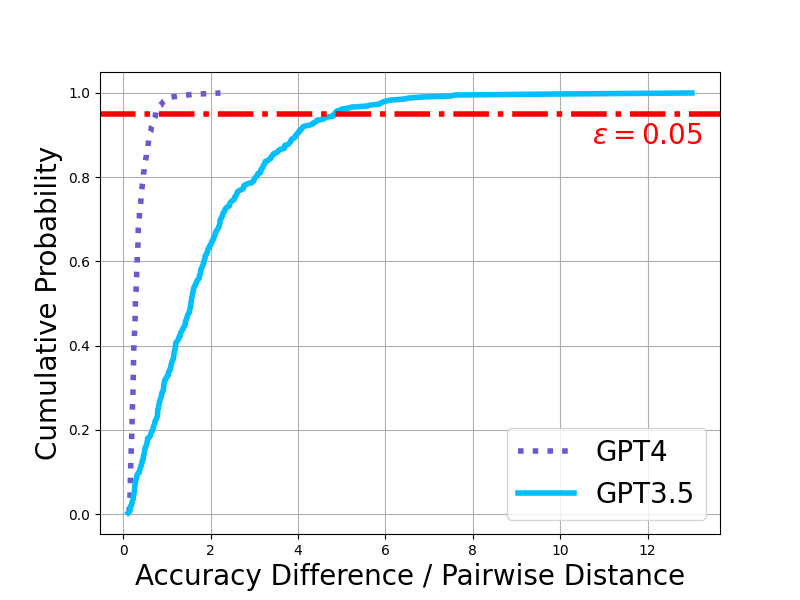}
    \end{subfigure}
    \begin{subfigure}[b]{0.325\textwidth}
        \centering
        \includegraphics[width = \linewidth]{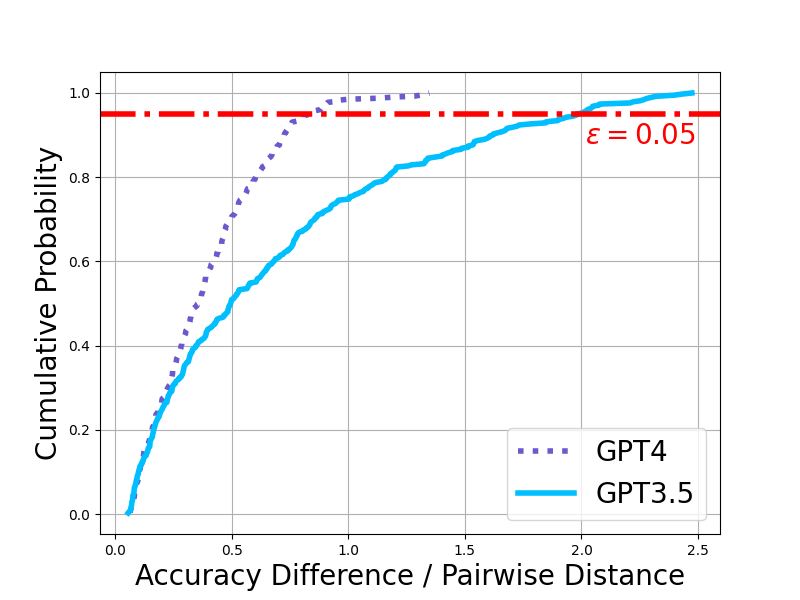}
    \end{subfigure}
      \begin{subfigure}[b]{0.325\textwidth}
        \centering
        \includegraphics[width = \linewidth]{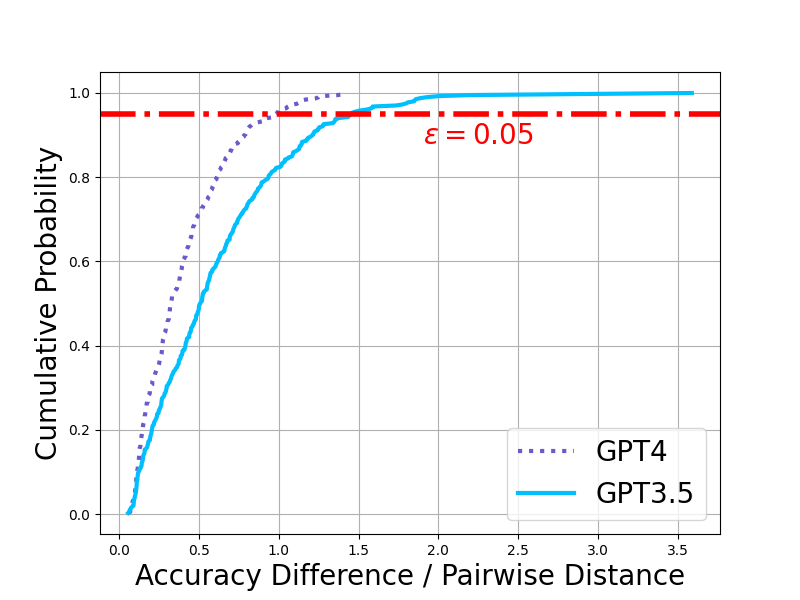}
    \end{subfigure}
    \caption{Estimating (probabilistic) Lipschitz constant of models (Definition \ref{def:k_delta_lipschitz}) on (left) Ethos (middle) GSM8K and (right) MedQA datasets for GPT-4 and GPT-3.5 models.}
\label{fig:motivation}    
\end{figure*}

Consider the class of sequential algorithms like ProTeGi \citep{pryzant2023automatic} and TextGrad \citep{houtextgrad} . 
The objective is to improve the accuracy of a given (black-box) \textit{solver LLM} $f:\mathcal{X} \rightarrow \mathbb{R}$ that takes as input a prompt $\mathbf{x} \in \mathcal{X}$ and outputs the average accuracy on a validation set $D_v$. 
Since the set of prompts is combinatorially large, we assume that all prompts can be embedded in a vector space such that distance between two prompts in the space correspond to their semantic similarity. 
The prompt optimization problem can be written as $\arg \max_{\mathbf{x} \in \mathcal{X}} f(\mathbf{x}; D_v)$.



Previous work has shown that LLMs can be  brittle to their input: changing the prompt slightly can create a significant difference in performance~\citep{zhuo2023robustness}.
We want to understand if the optimization problem is well-conditioned. Typically, conditioning can  be determined by the Hessian. However, since $f$ is black-box, we approximate it by measuring sensitivity, or more specifically, Lipschitz continuity near the optimal solution. Based on prior work on defining continuity of neural networks~\citep{nori}, we use a probabilistic notion. 

\begin{definition}[Probabilistic Lipschitz Continuity~\citep{nori}]
\label{def:k_delta_lipschitz}
Given a probability distribution over inputs $\mathcal{X}$, $r \geq 0$, and a distance measure $d$ such as $\ell_1$ or $\ell_2$ norm, a function $f: \mathcal{X} \rightarrow \mathbb{R}$ is $(L, \epsilon)$-probabilistically Lipschitz with constant $L \geq 0$, if
\begin{align}
    \Pr_{\mathbf{x},\mathbf{x}' \sim \mathcal{X}}[&\operatorname{d}(f(\mathbf{x}),f(\mathbf{x}')) \leq L \cdot \operatorname{d}(\mathbf{x},\mathbf{x}') \nonumber \\
    &|  \ \operatorname{d}(\mathbf{x},\mathbf{x}') \leq r] \geq 1-\epsilon .
\end{align}
\end{definition}

Note the focus on small changes in input through the parameter $r$. Intuitively, the Lipschitz property bounds the maximum change in $f$ given a small change in input prompt. Typically, the lower bound of error for any sequential optimization algorithm over $f$ is directly proportional to the Lipschitz constant $L$~\citep{malherbe2017global}. Therefore, for faster convergence, it is desirable to have a low $L$, especially near the optimal solution. 

Empirically, we estimate $L$ by sampling task-relevant prompts so that they are close to the optimal solution. Then we make small changes to the prompt such that the semantic meaning stays the same and measure the change in $f$ (See Appendix \ref{app:sec2.1} for experimental details). We show the change in $f$ per change in input for GPT4 and GPT3.5 models in Figure~\ref{fig:motivation} for the Ethos, GSM8K and MedQA datasets. 
Assuming $\epsilon=0.05$, probabilistic Lipschitz constant $L$ for GPT4 is $<1$, whereas it is higher for GPT3.5. Thus, as the model sizes increases, the probabilistic Lipschitz constant decreases. So, \textbf{larger models are more amenable to prompt optimization.}

\subsection{Details on estimation of Lipschitz constant $L$}
\label{app:sec2.1}
TO calculate the Lipschitz constant for a given LLM and task, we take a human written promp and generate it's paraphrases using GPT-4. We prompt GPT-4 with the following text: ``\textit{You are given a sentence, you have to generate 30 paraphrases of the sentence, make sure that the core content of each paraphrase is same, you can use add, subtract or change words}". These paraphrases are then evaluated on the validation set $D_v$. For a measure of distance between two prompts, we take the cosine similarity between the embeddings of two prompts. We use \texttt{text-ada-002} for generating the text embeddings for prompts. 


\subsection{Computational Complexity}
\label{sec:complexity}
We now consider the compute complexity of \shortname{} in terms of the number of expert or solver LLM calls per epoch, stage-wise.  \\ \textbf{Clustering}: First, we evaluate all the training examples using the current prompt. Second, for every wrongly predicted example, we obtain feedback from the expert LLM. Third, for the given set of feedbacks, we use a single call to cluster it into $l$ clusters. Each of the above steps incurs $O(N)$ queries, so the total query complexity of the clustering stage is $O(N)$. Finally, for each example, i.e., (question, answer) pair, we simply map it to the $l$ clusters (no LLM calls). 
\\ \textbf{Mini-batch feedback and Batch-level aggregation}: At a given epoch, we evaluate every question in the mini-batch using the current prompt and the solver LLM (\textbf{$N$} queries overall). Next, we obtain one feedback over all the wrong questions in the mini-batch $m$ (\textbf{$N/|m|$} queries). We use one call to aggregate these feedbacks. For prompt selection, we evaluate $4$ prompts on the batch $b$ (2 per beam), so \textbf{$O(4|b|)$} queries per batch. Hence overall query complexity is $N + N/|m| + 4N + 1$ or $O(N)$. \\ With LLM throughput of 0.5 qps, a training + validation set of 300 examples, 10 clusters, and 20 epochs, it takes under 7 hours to train.

\subsection{SLM Training Details}
\label{app:SLM_train_details}
To induce the ability of structured prompt generation in a smaller language model, we curate a section-wise dataset of around 12,000 task-prompt pairs. The tasks for training dataset creation were taken from tasksource library~\citep{tasksource} that contains around five hundred classification tasks. We extract the task description from tasksource-instruct, which contains tasksource dataset recasted with instructions. For instance, the task description for BIG-bench Entailed Polarity task is, \textit{"Given a fact, answer the following question with a yes or a no"}.
The dataset provides diverse tasks and their short description, but not the human-generated prompts for each task. To approximate human-generated prompts, we use GPT-4 as a teacher model. 

By prompting GPT-4 with the task description and section description, we ask it to generate the contents of the section. To ensure that the generated section-wise prompts are concise and relevant, we prompt GPT-4 to not generate more than five lines of content for each section. We use  LLAMA2-13B model, which we finetune  using LoRA adapters as the auxiliary LM that generates sections. 

\subsection{Data Set Creation of Real-World Task}
\label{app:dataset_real_world_task}
We sample real user queries from a proprietary application, rewrite them using ML models, and ask expert judges to label the query-pairs as identical or otherwise based on prescribed guidelines. We use a set of 200 examples as training data, and an additional 50 examples as validation set, to learn a prompt using \uniprompt{}, starting from the one-line description: \textit{Tell if query A and query B have same intent or not}. The dataset is heavily biased towards positive samples, so the metric of success is improvement in accuracy, over the best manually-engineered prompt, on the positive and negative classes individually. For testing, we use a separate labelled set of 2527 examples from two geographies --- one where the training data was sampled from, and the other unseen.

\subsection{Prompt to Llama2-13B for fine-tuning}
\begin{quote}
\begin{verbatim}
### Instruction:
You are a prompt engineer, you have 
to write a structured prompt.
For the given task description, 
examples and section description, 
write the contents of the section 
that align with 
section description.

### Task Description:
{data_point['task_description']}

### Section Description:
{data_point['section']}:
{section_descriptions\
[data_point['section']]}

### Response:
{data_point['prompt']}

\end{verbatim}
\end{quote}

\subsection{Prompt Initialization}
\label{app:initalization_example}

\textbf{One line task descriptions:}
\begin{enumerate}
    \item Ethos: In this task, you have to determine whether a given text is hate speech or not.
    \item ARC: You have to solve the following science question.
    \item GSM8K: In this task, you are given a math question. You have to solve the question.
    \item MedQA: In this task, you are given a medical question. You have to solve the question.
\end{enumerate}

\onecolumn

\subsection{Additional Results} 
\label{additional_results}
\begin{table}[ht]
    \centering
        \caption{Performance (\%  solved problems) of \uniprompt{} (GPT-4-Turbo solver) on code generation datasets, compared to GPT-4 \citep{openai2023gpt4} and newer models.}
        \begin{tabular}{l|r|r}
          \toprule
          Method  & HumanEval & MBPP \\
          \toprule
          GPT-4 & 67.0 & 87.5 \\
          GPT-4-Turbo & 87.1 & 90.9 \\
          GPT-4o & 90.2 & 92.4 \\
          \uniprompt{} & \textbf{93.8} & \textbf{92.5} \\
          \hline 
          \toprule
        \end{tabular}%
        \label{tab:code}%
\end{table}
\begin{table}[ht]
  \centering
      \caption{Ablation of LLM choices for \uniprompt{} on the Ethos dataset.  `Init' and `Final' denote initial (i.e., task description) and final prompt accuracies.}
    \begin{tabular}{l|l|r|r}
    \toprule
      Expert LLM    & Solver LLM & Init & Final  \\
    \toprule
    GPT-3.5-T & GPT-3.5-T & 76.8 & 82.4  \\
    GPT-4 & GPT-3.5-T & 76.8 & 92.3  \\
    GPT-3.5-T & GPT-4 & 89.8 & 91.4  \\
    GPT-4 & GPT-4 & 89.8 & \textbf{94.3}  \\
     \hline 
    \toprule
    \end{tabular}%
  \label{tab:expertchoice}%
\end{table}

\begin{table*}[ht]
  \centering
   \caption{Ablation of design choices in \shortname{} with GPT-3.5-Turbo as the solver model.}
  \small
    \begin{tabular}{l|c|c|c|c}
    \hline
          & Ethos & ARC & MedQA & GSM8K \\
    \hline
    \shortname{} $-$ History   & 88.0  & 84.6  & 55.3  & 80.8 \\
    \shortname{} $-$ Clustering & 77.5  & 82.0  & 54.1  & 81.5 \\
    \shortname{} & 92.3 & 86.0  & 57.1  &  82.4 \\
    \shortname{} $+$ Greedy & \textbf{93.7}  & 90.5  & 55.5  &82.3 \\
    \shortname{} $+$ Fb Clustering &87.2&\textbf{91.2}& \textbf{58.3}& \textbf{82.5}\\
    
    \hline
    \end{tabular}%
   
  \label{tab:ablation}%
\end{table*}

\begin{table*}[ht]
  \centering
      \caption{Ablation on the initial prompt for \uniprompt{} (\textbf{best} test accuracy in bold).}
    \begin{tabular}{l|r|r|r|r}
    \toprule
      Init Prompt    & Ethos & ARC & MedQA & GSM8K \\
    \toprule
      Expert Prompt & 84.0 & 86.0 & 52.3 & \textbf{82.4} \\
    Llama Prompt & 92.0 & \textbf{90.5} & 55.5 & 81.5\\
    Task Description & \textbf{92.3} & 86.0  & \textbf{57.1}  & \textbf{82.4} \\    
     \hline 
    \toprule
    \end{tabular}%
  \label{tab:initprompt}%
\end{table*}

\begin{table*}[ht]
  \centering
  \caption{Comparison of \shortname{} (``Ours'') with MedPrompt, with GPT-4 as the solver model.}
  \footnotesize
    \begin{tabular}{l|c|c|c|c}
    \hline
          & MedQA & PubMedQA & MedMCQA & MMLU MG\\
    \hline
    Ours & 80.9 & 70.3 & 79.2 & 78.0\\
    Ours + kNN & 81.0 & 72.2 & 81.4 & 94.0\\
    Ours + kNN + CoT & 83.9 & 74.7 & 82.6 & 96.0\\
        Ours + kNN+ CoT + Ensemble &  \textbf{87.0} &  \textbf{75.6} &  \textbf{84.5} &  \textbf{99.0}\\
    MedPrompt & 80.6 & 71.2 & 79.1 & 98.0\\

    \hline
    \end{tabular}%
    
  \label{tab:medprompt}%
\end{table*}

 

\textbf{An example of sectioned initialization prompt generated using finetuned Llama Model}
\begin{verbatim}
Introduction:
Assume the role of a science expert and answer the given question by selecting one 
of the options A, B, C or D.
     1. Understand and solve science questions by selecting the best answer 
     from a given list of options.
     2. Identify the logic behind the choices provided and make an informed decision.
     3. Use contextual clues to choose the most accurate answer.
     4. Be aware of the differences between science and everyday language.

Task Description:
          Scientific inquiry: Science is the systematic study of
          the structure and behavior of the physical and natural
          world through observation and experiment. The 
          scientific method is a process for acquiring knowledge
          that has been improved upon since its inception in the
          17th century. It involves making observations, 
          formulating hypotheses as to their causes, and 
          experimenting with them to support or refute the 
          hypotheses.

Real-life Application:
1. Assisting Students in Science Classes:
    In the context of science education, the ability to solve
    science questions can help students to better understand and
    internalize the concepts. By familiarizing themselves with 
    the basic principles of science, students can develop a 
    stronger foundation of knowledge.
2. Improving Scientific Literacy:
    Scientific literacy is a critical skill in today's world, 
    where scientific knowledge is increasingly important. By
    solving science questions, individuals can improve their 
    understanding of scientific concepts and be more informed 
    about scientific developments.
3. Scientific Questions: 
    In daily life, there are many questions that require 
    scientific knowledge to answer. For example, understanding 
    the science behind certain phenomena, such as why a magnet 
    sticks to a refrigerator door, can help us in our day-to-day
    life.
4. Increased Awareness: 
    By answering scientific questions, we can develop a deeper 
    understanding of the world around us and increase our 
    awareness of scientific phenomena. This can help us in our
    daily lives and make us more knowledgeable individuals.
    

Background Knowledge:
    1. Understanding of the basic concepts of science and 
    physics, such as the difference between heat, temperature
    and friction.
    2. Basic knowledge of the different types of skin surfaces,
    such as dry, wet, rough, smooth, etc.
    3. Familiarity with the different types of magnets and their
    properties.
    4. Understanding of the different factors that affect the 
    adhesion of magnets to different surfaces.
    5. Knowledge of the different types of sedimentary rocks and
    their properties.

Challenges:
1. Ambiguity in the question:
    The question might be ambiguous in nature, and it can be 
    difficult to understand the exact meaning of the question.
    In such cases, it is important to read the question 
    carefully and identify the key concepts or keywords. This 
    can help in arriving at the correct answer.
2. Scientific terms or concepts:
    The question might contain scientific terms or concepts that
    are unfamiliar to the user. In such cases, it is important 
    to understand the meaning of these terms or concepts and 
    their relationship with the question.
3. Difficulty in understanding the question: 
    Sometimes, the question might be complex or abstract, making 
    it difficult to understand or interpret.
4. Misleading statements or information: 
    The question might contain misleading or false information, 
    making it difficult to determine the correct answer.
5. Contradiction: 
    The answer can be in conflict with well-known scientific 
    facts or principles. In such cases, it is important to make 
    a careful analysis of the evidence and choose the answer 
    that is most consistent with the available

Simplification:
1. Identify the key elements in the question:
    Ask yourself, "What is the main question in the question?"
    Identify the key elements and focus on them to solve the 
    problem.
2. Understand the context:
    Understand the context of the question and the background 
    knowledge you need to answer it.
3. Identify the answer choice:
    Identify the answer choice that best fits the context and 
    background knowledge.
4. Eliminate the distractors:
    Eliminate the distractors that don't fit

Tricks:
1. Read the question carefully: Understand the question and its
context. This will help in understanding the information and 
concepts needed to solve the question.
2. Identify the key concepts: Identify the key concepts and 
keywords in the question. This will help in understanding the 
main idea and focus on the relevant information.
3. Understand the question structure: Understand the structure
of the question. This will help in identifying the appropriate
answer option and avoiding distractions.
4. Look for clues: Look for clues in the question and the 
answer options
\end{verbatim}

\subsubsection{Examples of training dataset}
\begin{verbatim}

Task: super glue boolq
Task Description: With no explanation, label the following with
either False or True.
Section: description
Prompt:  As a prompt engineer, your task is to create a prompt 
that accurately labels statements as either True or False 
without providing any explanations. The statements may involve 
general knowledge, trivia, or facts about events, places, 
people, or objects. Ensure you have a strong understanding of 
the subject matter and write a concise and specific task 
description related to the given statements.

===================================================

Task: glue qnli

Task Description: With no explanation, label A to B with either 
entailment or not entailment

Section: background

Prompt: 1. Entailment means that the information in statement B 
can be inferred directly from statement A.

2. Not entailment means that the information in statement B 
cannot be inferred directly from statement A or is unrelated.

3. Understanding the context and relationship between the two 
statements is crucial for accurate classification.

==================================================

Task: bigbench undo permutation

Task Description: In the following sentences with shuffled 
words, find the correct order of word swaps to unscramble the 
sentence.

Section: tricks

Prompt: 1. Identify the key words or phrases in the task to 
understand the context of the sentence. Look for nouns, verbs, 
and adjectives that seem related or could logically fit together.

2. Start by solving the problem step by step and focus on one 
swap at a time. Breaking the problem into smaller sub-problems 
will make it easier to manage.

3. To make the task more manageable, first focus on swapping the 
words that are clearly out of place, such as words that should 
be at the beginning or end of the sentence.

\end{verbatim}

\subsection{Prompt for identifying important facets}
\label{app:gpt4_facet_prompt}
\begin{verbatim}
you are given a task, along with it's description, some examples
of how to solve the task and section descriptions. 
What do you think would be the most important sections to 
include for the given task. 
## Task
{task}
## Task Descirption
{tas_description}
## Examples
{Examples_string}
## Section Descriptions
{sections}
\end{verbatim}

\subsection{Clustering Type 1}
\label{app:clustering_prompts_1}
\begin{verbatim}
You are given a science question, you need to tell which broad 
topic is this question from. 
Question: {train_questions_new[ij]}
Answer: {answer}
Give your answer as a single word, between <Answer></Answer> 
tags like: <Answer>Thermodynmics</Answer> or 
<Answer>Botany</Answer>.
Subtopic: 

\end{verbatim}

\subsection{Clustering Type 2}
\label{app:clustering_prompts_2}
\begin{verbatim}
You are given a set of feedbacks, you need to cluster them into 
five groups based on similarity, and then provide a summary of 
each group. You can use the following feedbacks to cluster: \n 
{feedback}

provide each cluster explnation within the following tags: 
<Cluster></Cluster>



You are given a feedback and a set of clusters, you need to tell 
which cluster this feedback belongs to. 

The clusters are: \n {string_of_clusters}

The feedback is: {feedback}

give your final answer as the number of the correct cluster 
between <Answer></Answer> tags like: <Answer>1</Answer>.'''
\end{verbatim}

\subsection{Feedback Prompts}
\label{app:feedback_prompts}
\textbf{Feedback over mini-batch}
\begin{verbatim}
You are a teacher and you have to give feedback to your 
students on their answers. 

You are teaching how to solve math problems to your students. 
You are given a question, it's true answer and answer given by 
student. You are also given the explanations written by your 
students while solving the questions. 

The questions are answered wrong by the students. 
You have to tell why is the solution wrong and what information
is can be added to the in the Background Knowledge part that 
would have helped the student to write better explanations.

## IMPORTANT: You are also given a history of changes you made
to the background knowledge part and the change in student's 
accuracy after making the change. You have to use this history 
to make your feedback.

Be explicit and tell the exact information that can be added 
without further modification / addition. 

### IMPORTANT: Give feedback in form of instructions like  add a
section, add a subsection, set the content of a section, set the
content of a subsection, delete a section or delete a subsection
in the background knowledge part.

Give very granular feedbacks, like if the student has made a 
mistake in the calculation, then tell what is the mistake in the 
calculation and how to correct it, if the student has made a 
mistake in the concept, then tell what is the mistake in the 
concept and how to correct it. 

## Background Knowledge
     {current_prompt}

## History
    {history_string}


Now, it is your turn to give feedbacks to the students.
You can only provide a one line feedback.
\end{verbatim}
========================================\\
\textbf{Feedback over batch}
\begin{verbatim}
You are given a set of feedbacks for some problems. The set 
feedbacks for each problem separated by =========== symbol. 
You have to summarize the feedbacks into a final feedback. 
You are also given a set of wrong questions. You need to tell 
which edit can be applied to aid the student in solving the 
wrong question.

To achieve your task, try to follow the following steps;
1. Identify the general problem that is being solved by all the
feedbacks. 
2. Once you have identified the problem, try to make a new 
feedback that covers most of the 
feedbacks given. 
Let's say the problem in the first feedback is the absence of 
methods to solve linear equation and in the second feedback it
is the method to inverse a matrix.
You know that both of these problems can be caused by adding how
to solve convert a matrix  into row rediced echolon form. So, 
add that. 
3. Try and validate your feedback. Once, you have a feedback try
to see if it covers every 
feedback, if it does not cover any feedback, add that to your
new feedback.
4. See the wrong questions and try to identify what is the 
problem in the question. 
If the problem is not covered by your feedback, add that to your
feedback.
5. You can add specifics like examples, definitions etc make 
sure that the feedback is enough to be directly added without 
any modification.

You may use the following function templates-

add_section(sectioname)
add_subsection(section_name, subsection_name)
set_section_content(section_name, new_content)
set_subsection_content(section_name, subsection_name, new_content)
delete_section(section_name)
delete_subsection(section_name, subsection_name)

Your summary cannot include more than four functions.
Make sure that the content is useful, 
not just a very general statement. Something specific.

Instructions:
{edits}

Wrong Questions:
{wrong_examples_string}

Summary:
\end{verbatim}

\subsection{Editing Prompt}
\label{app:editing_prompts}
\begin{verbatim}
You are given an input prompt and a feedback, you have to 
incorporate the feedback into the input prompt and output the 
final prompt.
An example of the task is given below

### Input Prompt
Introduction: In this task you have to answer the given question.

### Feedback
The background knowledge is incomplete, it does not include what
are the factors that affect the water usage and how many water 
sources are there.
\\add_subsection("Background Knowledge")
\\add_subsection_content(water usage depends on the population, 
climate, economic development, and availability of water 
sources. There are two sources of water, surface water and 
groundwater.)

### Final Prompt
Introduction: In this task you have to answer the given question.
Background Knowledge: water usage depends on the population, 
climate, economic development, and availability of water 
sources. There are two sources of water, surface water and 
groundwater.

Only output the final prompt nothing else. 

### INPUT PROMPT
{current_prompt}

### FEEDBACK
{edits}


### FINAL PROMPT
\end{verbatim}

\subsection{Example of prompt evolution using our method}
\begin{figure*}[h]
    \centering
    \includegraphics[width = \textwidth]{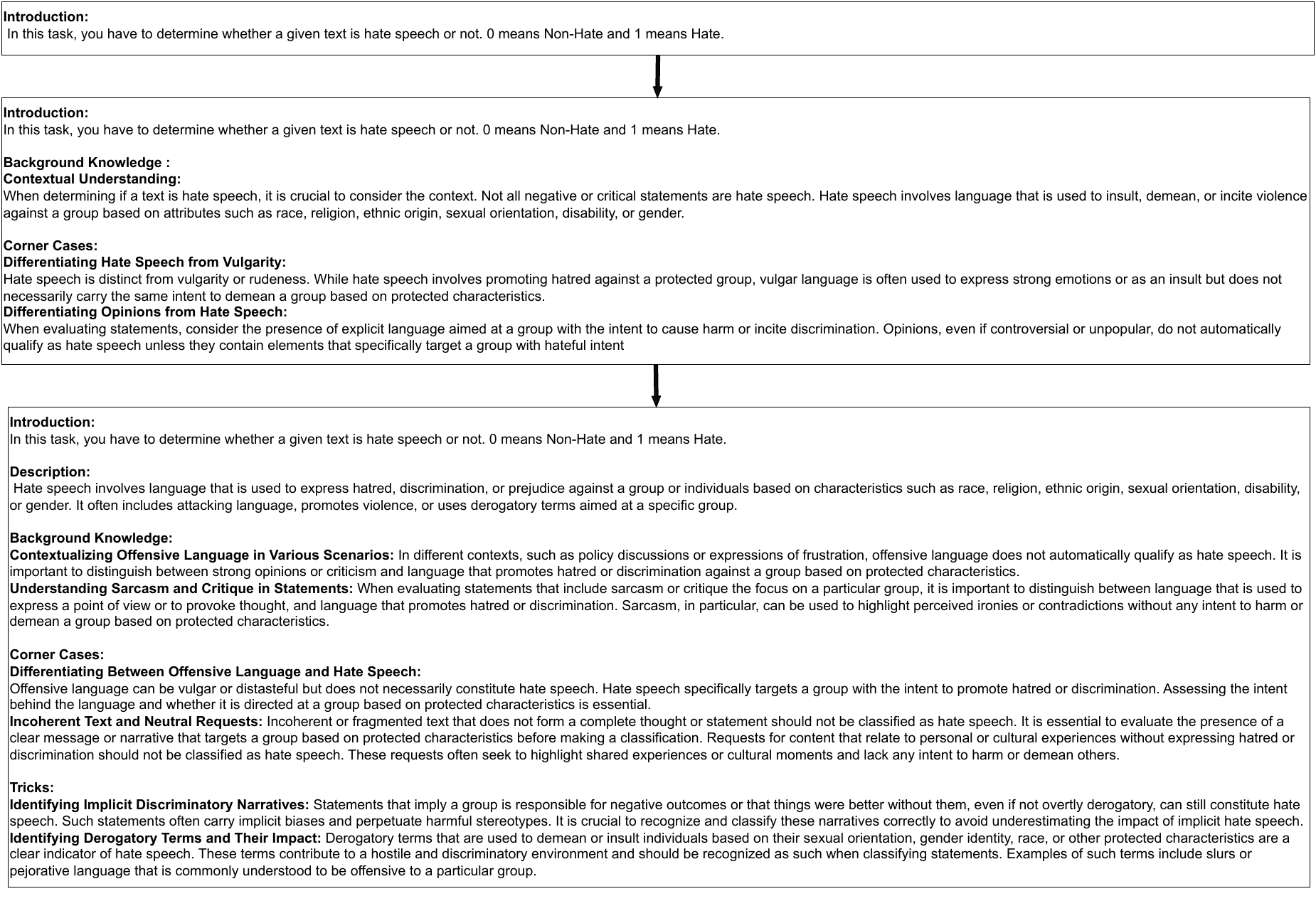}
    \caption{Evolution of prompts through iterations of \shortname{} on the Ethos dataset. Starting from a simple one-line prompt having an accuracy of \(82\%\), \shortname{} adds background knowledge, corner cases, and additional sub-sections yielding a prompt with accuracy \(88\%\). After further iterations, our algorithm converges to a detailed, human-like longform prompt that achieves accuracy of \(92\%\).}
    \label{fig:promptevo}
\end{figure*}

See example in Figure \ref{fig:promptevo}.

\subsection{Comparision of our method with existing methods}
See Figure \ref{fig:medqa}.

\begin{figure*}[!h]
\footnotesize
    \centering
    \begin{tcolorbox}
        \textbf{Human Prompt}\\
        \texttt{Let's differentiate using step by step reasoning like a medical expert.}\\
        \textbf{Our Prompt}\\
        \texttt{\textbf{Introduction}: In this task, you are given a medical question. You have to solve the question.\\\textbf{Description}:  To solve medical questions effectively, it is important to understand various medical conditions, their progression, and associated clinical features. 
        \\
        \textbf{Background Knowledge}: 
        Differential Diagnosis of Subcutaneous Nodules:\\
        When evaluating subcutaneous nodules, consider mobility, consistency, and skin adherence. Epidermoid cysts are firm, non-tender, and the skin cannot be pinched over them. Lipomas are soft, mobile, and have pinchable skin.\\
        \textbf{Corner Cases}: 
        Antiretroviral Therapy Complications:\\ Doctor should be aware of the common side effects of antiretroviral drugs, with specific attention to the association between didanosine and pancreatitis, and the recommended management strategies, such as replacing didanosine with lamivudine.}
        
    \end{tcolorbox}
 \caption{Comparison of human-written Prompt and prompt produced by \shortname{} on MedQA dataset.}
 
    \label{fig:medqa}
\end{figure*}
\begin{figure*}[t]
\footnotesize
    \centering
        \begin{tcolorbox}
        \textbf{OPRO optimized prompt}\\
        \texttt{Start by dissecting the problem to highlight important numbers and their relations. Decide on the necessary mathematical
operations like addition, subtraction, multiplication, or division,
required for resolution. Implement these operations, keeping in
mind any units or conditions. Round off by ensuring your
solution fits the context of the problem to ensure accuracy}\\
        \textbf{Our Prompt}\\
        \texttt{\textbf{Introduction}: In this task, you are given a math question. You have to solve the question.\\
        \textbf{Strategies for Word Problems}:\\
        1. Understanding Word Problems: When solving word problems, it is crucial to read each sentence carefully and comprehend the time periods and quantities involved. Avoid incorrect multiplication or addition by paying close attention to whether a quantity remains constant over a period or changes. If a quantity is consistent, it does not need to be multiplied by the number of days or weeks unless the problem specifies otherwise.\\
        2. Calculating Averages: To calculate the average of a set of numbers, add all the numbers together and then divide by the number of items. In word problems, ensure you have the correct total before dividing by the number of periods, such as weeks, to find the average for each period.\\
        3. Understanding Past and Future Events in Word Problems: Distinguish between past and future events by identifying the starting and ending points. To calculate the time interval between two events, determine the direction of time from past to future and compute the interval accordingly. This understanding is essential when dealing with problems that ask for the time since a past event or until a future event.
     }
        
    \end{tcolorbox}
    
      \caption{Comparison of prompt produced by the state-of-the-art ORPO \citep{LLMO} and by \shortname{} on the GSM8K dataset.}
      \label{fig:gsm}
\end{figure*}



\subsection{Effect of length on performance of prompt}
\label{app:promptlength}
Here we answer the question: \textit{How much does only length contribute to \uniprompt{}'s success?}. To answer this, we replace the prompt with in-context examples of the same context length and compare the accuracies in Table \ref{tab:length}. We also compare the case where we include only the examples that the solver LLM gives incorrect prediction on, denoted as ``Wrong ICL'' row in the table. We see that there is a slight increase in accuracy when wrong examples are included in the prompt over randomly including examples. But, overall, \shortname{} performs much better than including in-context examples. This shows that length is not the only factor contributing to \uniprompt{}'s success.
\begin{table*}
    \centering
    \caption{Analysis of the effect of length and contents on the performance of \shortname{}}
    \begin{tabular}{l|ccc}  
    \hline
       & Ethos & ARC& GSM8K\\
       \hline
\shortname{} &93.7&90.5&82.4\\
ICL Prompt&63.0&86.7&76.3\\
Wrong ICL&70.4&87.1&78.2\\
Summarized Prompt&84.3&85.5&66.0\\ 
\hline
    \end{tabular}
    
    \label{tab:length}
\end{table*}

\subsection{Do diverse task facets organized as sections really help?}
\label{app:facetablation}
We want to empirically validate if all the diverse task facets that \uniprompt{} learns indeed contribute to the performance gains that we observe in Table \ref{tab:performance}. We consider two ablations: 

\textbf{1)} We successively remove each facet (i.e., sections) in the learnt prompt for the task and report the performances of the prompts with fewer facets. In Figure \ref{fig:trainingSoTA}, for the Ethos dataset,  we see that almost every additional facet contributes to non-trivial gains in accuracy. 

\textbf{2)} Could we have captured the information differently and retained the performance? We do a simple experiment -- we summarize all the facets (i.e., learnt prompt) and evaluate the resulting prompt. In Figure \ref{fig:trainingSoTA} (right) (green line), we see that the summarized prompt has a significant accuracy drop.   

\begin{table*}[htbp]
    \centering
    \caption{Sensitivity of \shortname{} to expert LLM prompts, on the Ethos dataset.}
    \begin{tabular}{c|r}  
    \hline
       Expert LLM Prompt for \shortname{} & Test Accuracy \\
       \hline
Simple prompt for mini-batch feedback &83.5\\
Simple prompt for batch feedback &91.0\\
Detailed prompts (Appendix \ref{app:feedback_prompts})  &93.7\\
\hline
    \end{tabular}
    \label{tab:promptsensitivity}
\end{table*}

\subsection{Sensitivity to prompts used for expert LLMs in \uniprompt}
\label{app:promptsensitivity}
The prompts used for expert LLMs in our algorithm, i.e., for clustering, feedback over batches and mini-batches, and editing, do matter for obtaining good performance. However, note that the prompts are task-agnostic and can be used as-is for new tasks. Moreover, prompts for clustering and editing are very simple and involved minimal human effort. Further, to study the reliance of \uniprompt{} on the quality of feedback prompts, we run an ablation study, where we replace the engineered prompts for feedback at batch and mini-batch levels with simpler prompts. The results are given in Table \ref{tab:promptsensitivity} for the Ethos dataset. We observe that the performance of \uniprompt{} depends heavily on the prompt used for obtaining feedback at mini-batch level; whereas simplifying prompt for feedback at the batch level has much less impact on the final accuracy.

\subsection{\shortname{} training behavior}
\label{app:trainingcurves}

An example of evolution of prompts using our algorithm is given in Appendix \ref{fig:promptevo}. It starts with a simple description of task and adds important facets like \textit{differentiating between hate speech and rudeness}. In contrast, \textbf{ProTeGi} \citep{pryzant2023automatic} yields a rather terse prompt on the same dataset: ``\textit{Does the following text contain language that targets a group of people based on their religion, gender, or other personal characteristics?}''. 

\begin{figure*}[htbp]
    \centering
    \begin{subfigure}[b]{0.66\textwidth}
    
         \includegraphics[width = 0.49\textwidth]{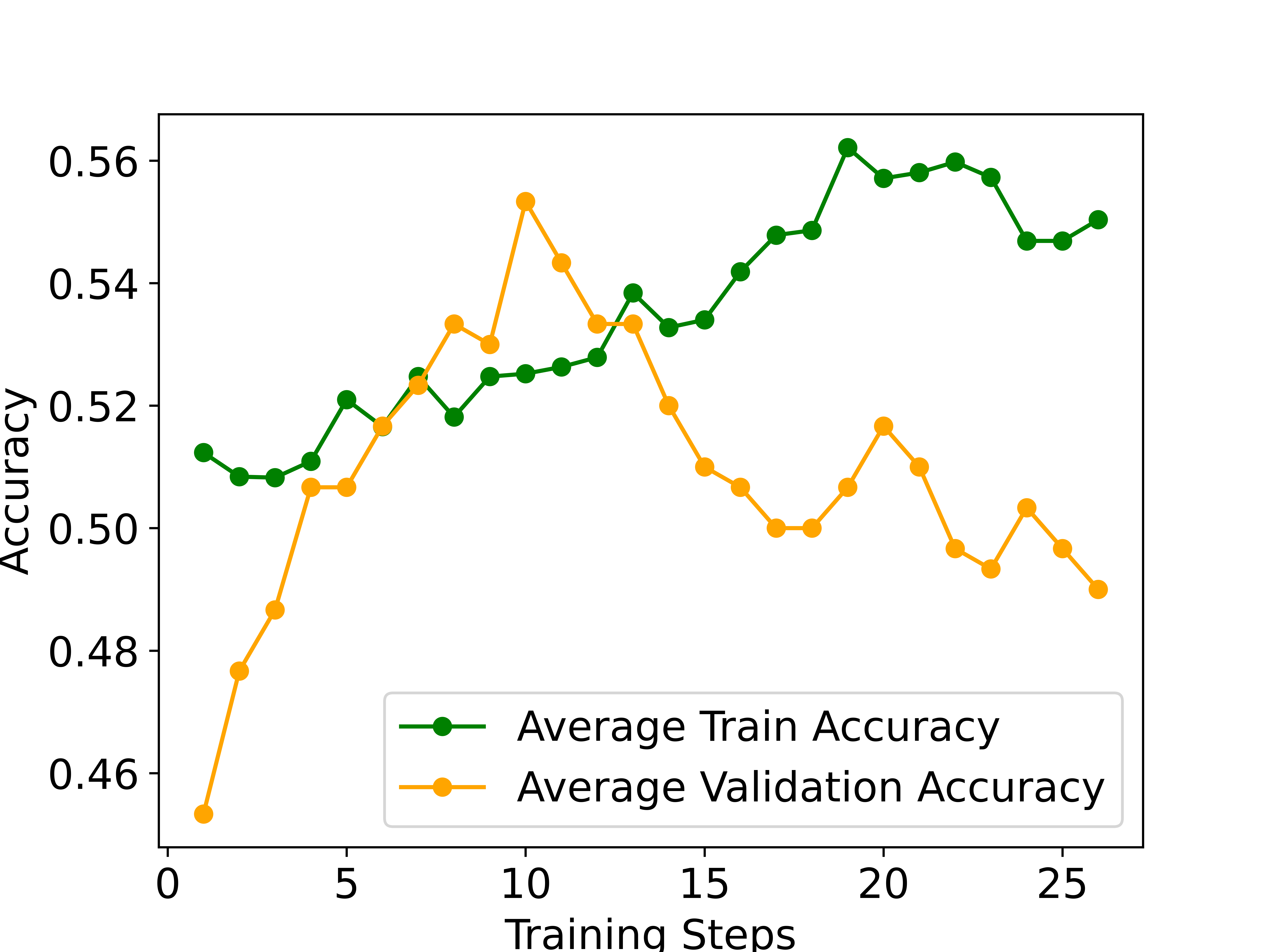} 
         \includegraphics[width = 0.49\textwidth]{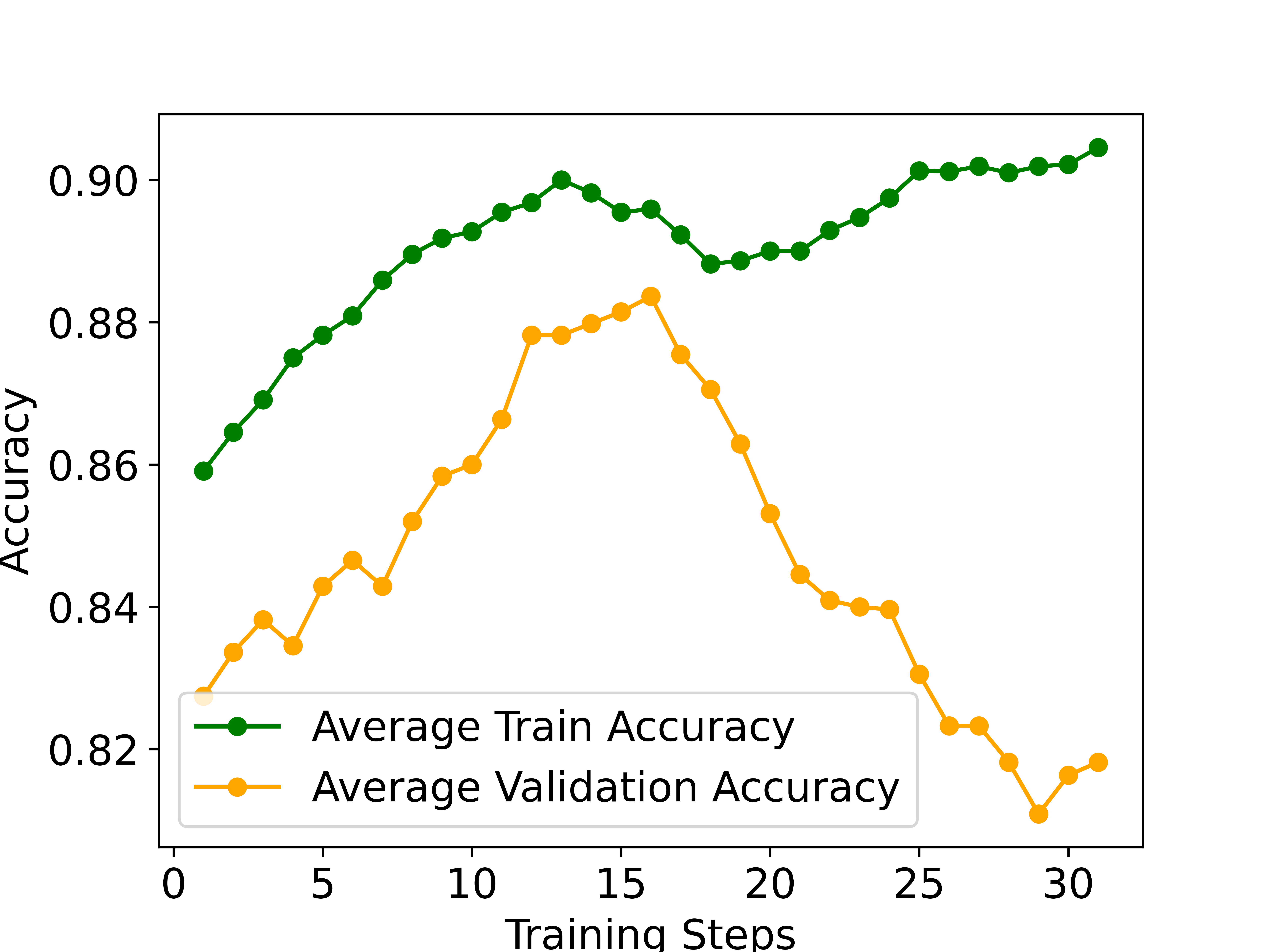}
        
    \end{subfigure}
    \begin{subfigure}[b]{0.33\textwidth}
    \includegraphics[width = \textwidth]{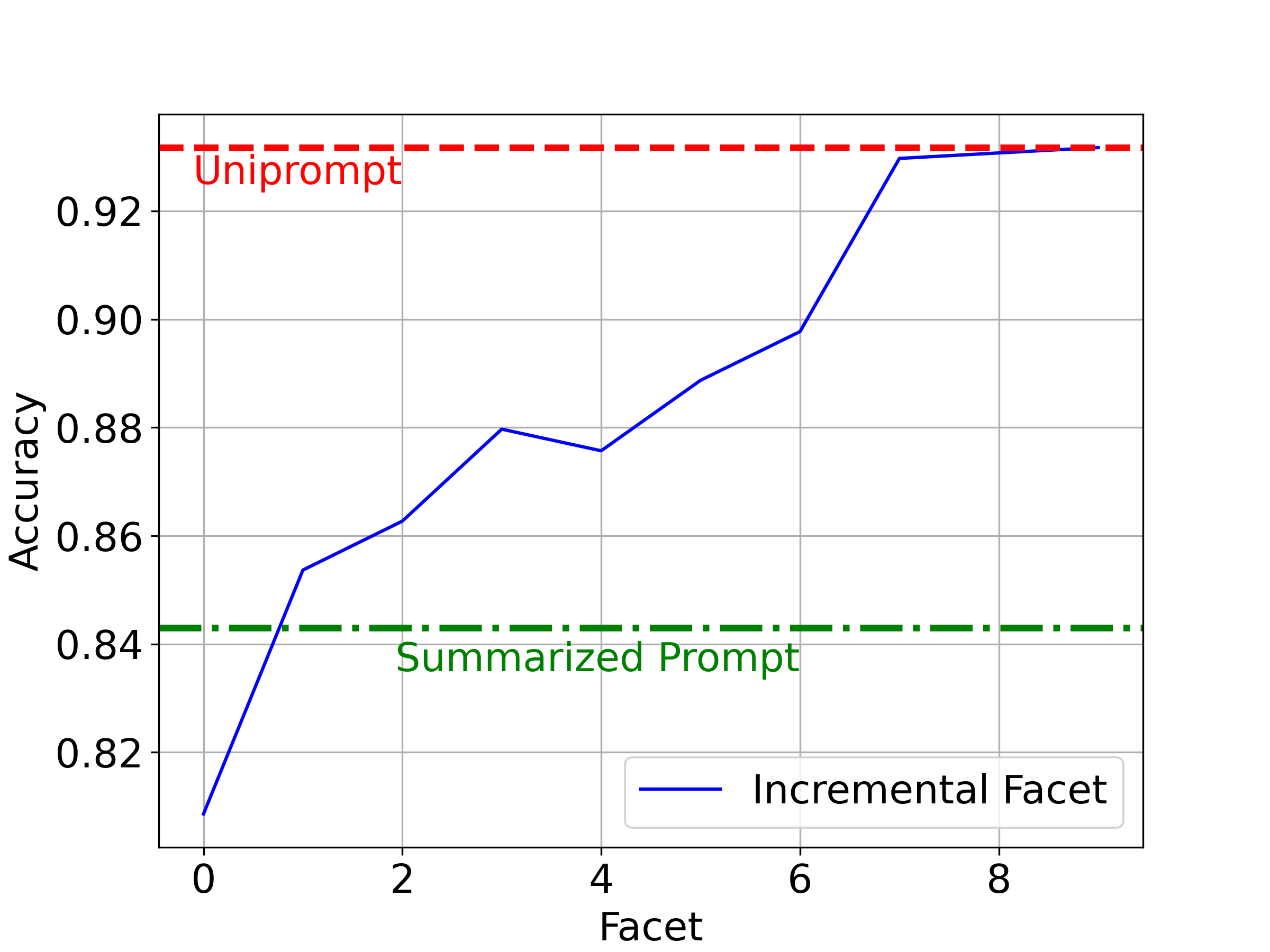}
    \end{subfigure}
    \caption{Training curves for MedQA (left) and ARC (middle) datasets when \uniprompt{} is initialized with (published) state-of-the-art prompts; (right) ablation of facets on Ethos.}
    \label{fig:trainingSoTA}
\end{figure*}

The training curves in Figure \ref{fig:trainingSoTA} show that our method initially performs edits on the prompt that simultaneously increase the train as well as the validation accuracy. After about 10 or 15 iterations (each batch update is an iteration), validation accuracy decreases while train accuracy continues increasing, indicating overfitting; which we overcome using early stopping.

\end{document}